\documentclass{amia}
\usepackage{pdflscape}
\usepackage{float}
\usepackage{multirow}
\usepackage{float}
\usepackage{booktabs}
\usepackage{amsmath}
\usepackage{graphicx}
\usepackage{amssymb}
\usepackage{mathrsfs}
\usepackage{tabularx}
\usepackage[table]{xcolor}
\usepackage{adjustbox}
\usepackage{tablefootnote}
\usepackage{subcaption} 
\usepackage{makecell}
\usepackage[labelfont=bf]{caption}
\usepackage[labelsep=period]{caption}
\usepackage{etoolbox}

\BeforeBeginEnvironment{align*}{\vspace{-0.5cm}}
\AfterEndEnvironment{align*}{\vspace{-0.5cm}}

\begin{document}

\title{The Effect of Enforcing Fairness on Reshaping Explanations in Machine Learning Models}

\author{Joshua W. Anderson, MS$^1$, Shyam Visweswaran, MD, PhD$^{1,2}$}

\institutes{
    $^1$ Intelligent Systems Program, University of Pittsburgh, Pittsburgh, PA\\
    $^2$ Department of Biomedical Informatics, University of Pittsburgh, Pittsburgh, PA 
}

\maketitle

\section*{Abstract}
\textit{Trustworthy machine learning in healthcare requires strong predictive performance, fairness, and explanations. While it is known that improving fairness can affect predictive performance, little is known about how fairness improvements influence explainability, an essential ingredient for clinical trust. Clinicians may hesitate to rely on a model whose explanations shift after fairness constraints are applied. In this study, we examine how enhancing fairness through bias mitigation techniques reshapes Shapley-based feature rankings. We quantify changes in feature importance rankings after applying fairness constraints across three datasets: pediatric urinary tract infection risk, direct anticoagulant bleeding risk, and recidivism risk. We also evaluate multiple model classes on the stability of Shapley-based rankings. We find that increasing model fairness across racial subgroups can significantly alter feature importance rankings, sometimes in different ways across groups. These results highlight the need to jointly consider accuracy, fairness, and explainability in model assessment rather than in isolation.}

\section*{Introduction}
Machine learning (ML) models are increasingly used in clinical settings to support diagnosis, risk stratification, and treatment decision-making. Their ability to learn complex patterns from high-dimensional data has made them valuable tools for improving efficiency and potentially enhancing patient outcomes. However, as these systems become more deeply integrated into clinical workflows, concerns about their reliability, transparency, and fairness have become central to discussions about their safe and responsible deployment \cite{goisauf2025trust}. Given the ethical complexities and inherent uncertainty in clinical practice, clinicians need to understand and validate ML recommendations before acting on them \cite{burkart2021survey}.

One of the most common approaches to transparency is the use of model explanations, which aim to visualize how data influences a model's reasoning, making predictions more understandable to clinicians, researchers, and other stakeholders \cite{hur2025comparison}. Inherently interpretable, white-box models can be understood through the structure itself as the primary explanation. For example, the coefficients of a linear model can explain the association between features and the predicted outcome. In contrast, many complex models (e.g., ensemble methods, gradient-boosted trees, deep networks) function as black-box systems whose internal mechanics are too complex to interpret directly. In these cases, post-hoc explanation methods provide an interpretable approximation of the original model's behavior \cite{shap}.

Alongside the need for transparency, there is growing concern about model fairness in clinical ML \cite{shaikh2022reassessment}. Fairness and bias are two terms that refer to the same fundamental issue: disparities in how models perform across different groups. Fairness metrics are used to identify and quantify these disparities in model performance, while bias mitigation strategies aim to reduce them. Improving fairness, or reducing bias, in ML models, particularly across racial groups, is crucial to minimizing unfair outcomes that disproportionately affect marginalized or historically underserved populations \cite{rabonato2025systematic}.

The intersection of model explanations and model fairness has been previously studied. Prior work has shown that explanation methods can support fairness evaluation and can also be incorporated into bias mitigation procedures \cite{begley2020explainability, ramachandranpillai2025fairxai}. Explanations have also been found to influence human judgments of fairness, highlighting their role in shaping perceptions of algorithmic systems \cite{zhou2020towards}. Prior work has largely focused on explanations as tools for achieving fairness. In this study, we evaluate how fairness constraints reshape explanations. We measure this interaction across three axes: predictive performance, fairness performance, and model explanation variability. These three dimensions capture not only how fairness constraints affect model accuracy and group-level fairness, but also how they alter the explanations that end users rely on to interpret model reasoning.

\section*{Background}

\textbf{Explainability}. SHAP offers several advantageous properties, such as local accuracy, missingness, and consistency, and provides a principled way to generate additive feature-importance scores for individual predictions \cite{lundberg2017unified}. Local explanations are provided by SHAP values, which quantify how much each feature shifts a single prediction above or below the model's baseline or average output. They represent how much each feature's value moved the model's output away from the baseline or average prediction for that specific instance \cite{trindade2024impacts}.

 Global explanations are generated by summarizing SHAP values across multiple predictions to assess each feature's overall influence. By examining how consistently a feature raises or lowers predictions and by how much on average, typically using the mean absolute SHAP value, one can assess its overall importance.

\begin{figure}[h!]
    \centering
    \begin{subfigure}[t]{0.49\textwidth}
        \centering
        \caption{Local Explanation}
        \vspace{7mm}  
        \includegraphics[width=\linewidth]{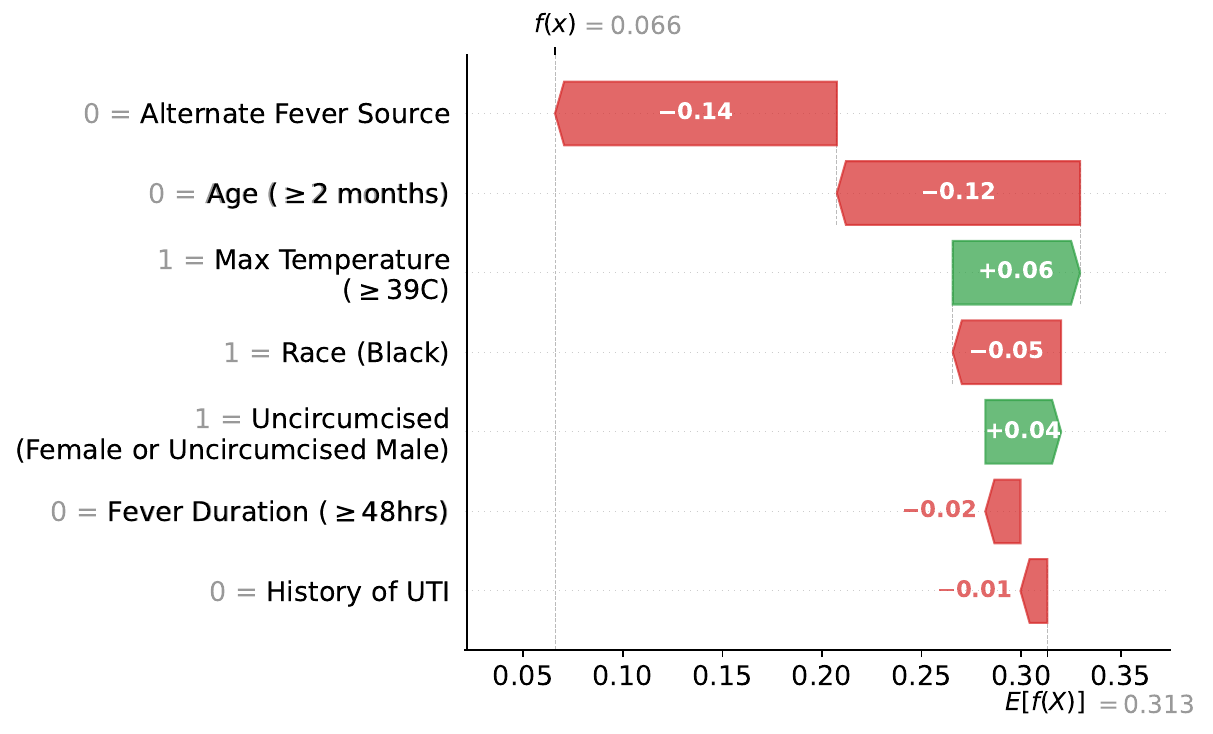}
        \label{fig:shapa}
    \end{subfigure}
    \hfill
    \begin{subfigure}[t]{0.49\textwidth}
        \centering
        \caption{Global Explanation}
        \vspace{7mm}  
        \includegraphics[width=\linewidth]{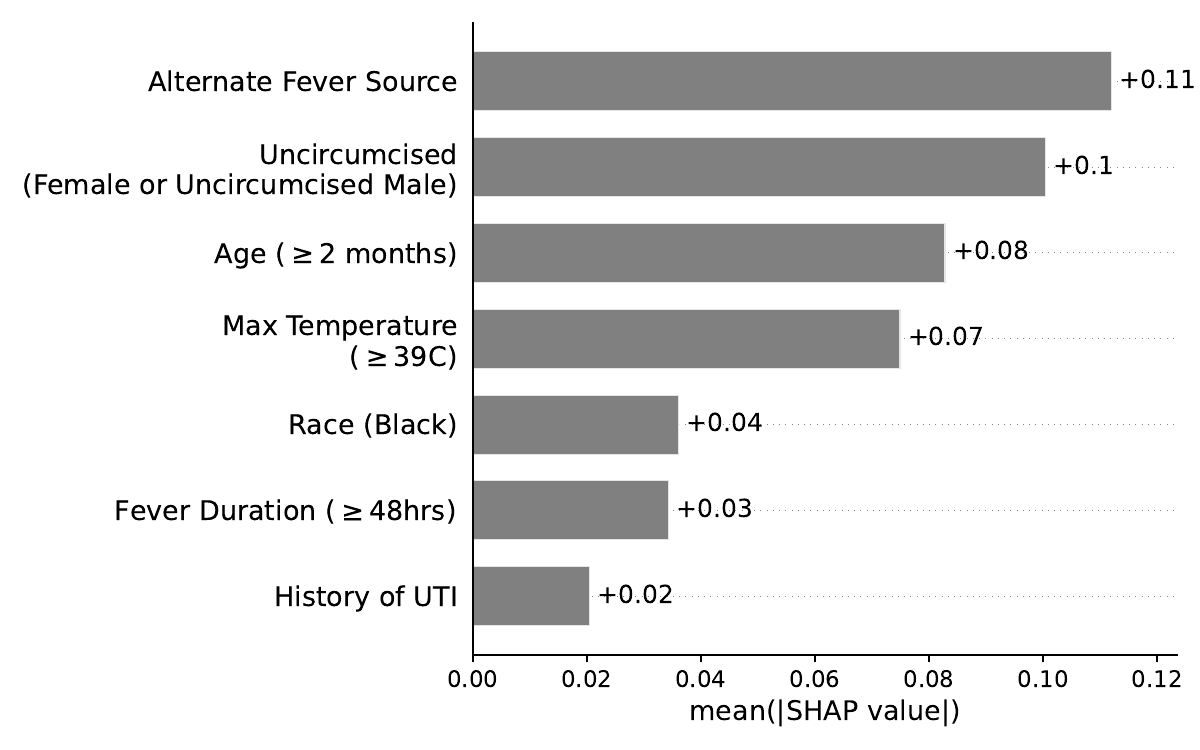}
        \label{fig:shapb}
    \end{subfigure}
    \caption{Example local and global explanations for predicting urinary tract infection (UTI) in pediatric patients. (a) Waterfall plot for a patient in the UTI dataset, showing SHAP feature importance values corresponding to the patient's predicted probability of having a UTI. (b) Bar plot showing the mean absolute SHAP values for each feature in the UTI dataset, demonstrating the ranking and relative influence of features on the model's predicted probability of UTI in pediatric patients.}
    \label{fig:sidebyside}
\end{figure}

Global explanations are obtained by aggregating SHAP values across predictions for many instances to determine each feature's overall influence. By examining how consistently a feature raises or lowers predictions and by how much on average, typically using the mean absolute SHAP value, one can assess its overall importance. Figure \ref{fig:sidebyside} shows examples of both (a) local and (b) global explanations for a model predicting risk of urinary tract infection (UTI) in children.

The computational complexity of $\phi(f,x)$ makes computing exact values impractical for large datasets, so several estimation methods have been developed to leverage model architectures for efficient computation. Lundberg and Lee proposed the most popular approximation methods, including KernelSHAP (model-agnostic), DeepSHAP (neural networks), and TreeSHAP (tree ensembles) \cite{lundberg2017unified, lundberg2018consistent}. 


\textbf{Fairness in Machine Learning}. The legal field distinguishes two core forms of discrimination: disparate treatment and disparate impact. Disparate treatment, called direct discrimination, occurs when individuals are explicitly treated differently because they are part of a protected group. Disparate impact, called indirect discrimination, occurs when a seemingly neutral regulation disproportionately harms members of a protected group. ML models trained without protected attributes that directly indicate group membership are less prone to causing disparate treatment. However, they can still unintentionally produce disparate impact and should be evaluated accordingly \cite{binns2020apparent}. Fairness is measured using protected attributes for which disparate performance or treatment may raise ethical or legal concerns, such as gender, age, socioeconomic status, and disability. There are three most common fairness metrics:

\begin{itemize}
    \item \textit{Demographic parity (DP; statistical parity)}: A model satisfies demographic parity if the overall rate of positive predictions is the same across protected groups (e.g., Black vs. non-Black groups).
    \item \textit{Equalized odds (EOD)}: A model satisfies equalized odds if true positive rates and false positive rates are equal across protected groups.
    \item \textit{Predictive parity (PP; predictive value parity)}: A model satisfies predictive parity if the positive predictive value (PPV), the probability that a positive prediction is correct, is the same across protected groups.
\end{itemize}

For most predictive models, whether in medicine or in domains like criminal justice, DP is rarely appropriate because of group-level differences in the predicted outcomes. For example, different racial groups have different recidivism rates, with Black individuals generally showing higher rates of re-offending compared to White individuals in the United States \cite{propublica_compas_2016}. Additionally, young children of different racial groups have different UTI rates, with Black children having higher rates of UTI compared to others. In screening models, which aim to avoid missed disease, the most appropriate metric is EOD, especially equal sensitivity. PP can be used as a secondary metrc, if positive results trigger invasive follow-up.

These fairness constraints are mathematically orthogonal, resulting in a phenomenon known as the impossibility theorem \cite{brcic2021impossibility}. For example, optimizing for equalized odds (EOD) may conflict with demographic parity (DP) in datasets where the base rates of positive outcomes differ between groups. Given protected attribute, $A$, prediction $\hat{Y}$, and outcome $Y$, the theorem states:

\begin{itemize}
    \item If $A$ and $Y$ are not independent, then DP and PP cannot simultaneously hold.
    \item If $A$ and $\hat{Y}$ are not independent of $Y$, then DP and EOD cannot simultaneously hold.
    \item If $A$ and $Y$ are not independent, then EOD and PP cannot simultaneously hold.
\end{itemize}

In addition to the impossibility of satisfying all fairness constraints, there is often a trade-off between enforcing fairness constraints and optimizing accuracy. However, it is debated if this trade-off is observed in all situations \cite{anderson2025algorithmic, cooper2021emergent}.

There are several ways to mitigate bias once a fairness metric is defined. Pre-processing methods, such as data reweighting or correlation removal, modify the input data to reduce the influence of protected attributes \cite{bird2020fairlearn}. In-processing methods, such as fair adversarial training, adjust the learning algorithm to penalize bias during model training \cite{xu2021robust}. Post-processing techniques, like threshold adjustments, modify the model's outputs to align predictions with fairness constraints \cite{lohia2019bias}. Additionally, other axes of fairness differ from this group-based notion in their underlying philosophy of the problem. Individual fairness proposes that "similar individuals be treated similarly" and typically involves pairing individuals through some similarity metric \cite{anderson2025algorithmic,dwork2012fairness}. Counterfactual fairness similarly considers individuals and proposes a causal view of fairness. Methods for counterfactual fairness consider if individuals would have been treated differently if they had belonged to another group in a counterfactual world \cite{anderson2025algorithmic, kusner2017counterfactual}. Each approach offers unique advantages and limitations, and the choice of mitigation strategy depends on the specific fairness goals, dataset characteristics, and domain requirements.

\section*{Methods}
\textbf{Datasets}. The UTI dataset, comprising young, febrile children admitted to UPMC Children's Hospital of Pittsburgh, was used to develop UTICalc, a tool to help clinicians assess UTI risk without requiring a urine sample \cite{shaikh2008prevalence}. The dataset consists of several binary demographic and clinical features from febrile children aged 2 years or younger: age ($<$12 months vs. $\geq$12 months), sex (female or uncircumcised male vs. circumcised male), race (Black vs. non-Black), fever ($<$39°C vs. $\geq$39°C), presence of an alternate fever source (no other source vs. other source), history of UTI (yes vs. no), and duration of fever ($<$48 hours vs. $\geq$48 hours). While an observed association between UTI and race initially motivated including race as a predictor in UTICalc, subsequent analysis revealed that this inclusion encoded racial bias, perpetuating existing disparities in treatment \cite{shaikh2022reassessment, anderson2024measuring, binns2020apparent}.

The Correctional Offender Management Profiling for Alternative Sanctions (COMPAS) risk assessment tool, used in U.S. courts to estimate a defendant's likelihood of recidivism (re-offending), is a widely used benchmark for fairness research. COMPAS was designed to calibrate risk scores (i.e., predictive parity), meaning that for any given score, the proportion who reoffend is similar across groups. And indeed, COMPAS was calibrated across races. However, the model exhibited racial disparities in error rates, even when overall predictive accuracy was similar across groups. 

The third dataset is a cohort of 24,468 patients with nonvalvular atrial fibrillation (AFib) treated with direct oral anticoagulants (DOAC) and managed by cardiologists within a single health system, aiming to predict major bleeding events requiring hospitalization within one year \cite{chaudhary2025machine}. Of the 2.3\% experiencing a major bleed, the patients were generally older and had more comorbidities, including hypertension, heart failure, coronary artery disease, anemia, and prior stroke. The dataset consists of electronic health record (EHR) variables, including multiple demographic variables such as race. 

Race variables are condensed to binary (1=Black, 0=non-Black) to measure fairness across race in a standardized way across different datasets. The UTICalc data was measured initially in this form and did not need to be modified. AFib and COMPAS data represented more granular measurements. For AFib, the value 'non-Black' represents values' Indian (Asian)', 'Vietnamese', 'Other Pacific Islander', 'Korean', 'Hawaiian', 'White', 'Japanese', 'Filipino', 'Alaska Native', 'Chinese', 'Guam/Chamorro', 'American Indian', 'Other Asian', 'Samoan', 'Declined', 'Other', and 'Not Specified' from the original data. For COMPAS, the value 'non-Black' represents values 'Caucasian', 'Asian', 'Hispanic', 'Native American', and 'Other' from the original data.

\textbf{Models}. The models used in this study represent a spectrum of ML complexity and functionality. Logistic regression is a linear model, ideal for tasks where the relationship between inputs and outputs is approximately linear, offering simplicity, interpretability, and computational efficiency. Random forests leverage multiple decision trees trained on random subsets of the samples and variables to improve accuracy and robustness, especially in datasets with nonlinear relationships. XGBoost (Extreme Gradient Boosting) is another ensemble model that builds trees sequentially and is particularly effective at handling structured data and achieving high performance. Neural networks are highly flexible models that can learn learn complex relationships in data at the cost of transparency. Each of these model types was chosen to add diversity to the results and capture how fairness mitigation may affect different architectures differently \cite{meyfroidt2009machine, chen2016xgboost}. Default probability cutoffs of 0.5 were used to ensure consistent comparisons of predictions across models.

\textbf{Feature importance}.
SHAP values are computed for baseline models and mitigated models. Since the computation of exact SHAP values is infeasible, we use approximation methods to find approximate SHAP values. The approximation methods we used included  KernelSHAP for logistic regression, TreeSHAP for random forests and XGBoost, and DeepSHAP for the neural network. We used the approximation methods implemented in Lundberg's SHAP library (v0.48.0).

\textbf{Fairness}.
To measure the fairness of each model, the EOD metric is used \cite{bird2020fairlearn}. This metric best generalizes across the three case studies to achieve their fairness goals. Given features race, $A$, prediction, $\hat{Y}$, and true outcome $Y$, EOD is defined as:

\begin{align*}
    EOD_{Difference} = P(\hat{Y} = 1| Y = y, A = Black) - P(\hat{Y} = 1 | Y=y, A = non-Black),\ y\in {0,1}
\end{align*}

Since EOD enforces parity in both the true positive rate ($y=1$) and the false positive rate ($y=0$), its overall value is the maximum of these two component metrics. We used Microsoft's Fairlearn (v0.13.0-dev) for bias mitigation to implement exponentiated gradient reduction, an in-processing method that runs constrained optimization using fairness metrics (e.g., EOD) as constraints \cite{bird2020fairlearn, agarwal2018reductions}. Once the mitigated models are trained, we compare their performance and explainability with those of the baseline models.

\textbf{Experiments}. For each of the three datasets, we analyze a baseline model and a mitigated model obtained via the exponentiated gradient reduction method. Bias mitigation uses accuracy for prediction optimization and EOD as a fairness constraint. We then report the predictive and fairness performance metrics for each model. Finally, we construct feature importance rankings for each model using global SHAP explanations and perform statistical tests to evaluate our hypothesis.

\textbf{Statistical Testing}. We compare changes in feature importance after bias mitigation by testing differences in importance ranking using Spearman's rank correlation coefficient ($\rho$). This test is a nonparametric measure of the monotonic relationship between two ranked variables. We measure importance ranking by taking the mean SHAP magnitude across all instances of a variable. By computing Spearman's $\rho$ between the pre- and post-mitigation feature importance, we obtain a single statistic that quantifies the degree of concordance in ranking: $\rho = 1$ indicates that the rankings are identical (no change), $\rho = 0$ indicates no correlation in ranks (complete reordering), and $\rho = -1$ indicates a perfect inverse ranking. This is used to test the following hypothesis:

\begin{align*}
H_0 &: \rho = 0 \quad \text{(no monotonic association between pre- and post-mitigation feature rankings)} \\
H_1 &: \rho \neq 0 \quad \text{(a monotonic association exists between pre- and post-mitigation rankings)}
\end{align*}

Since SHAP values are computed at the local explanation level, we perform subsequent tests conditioned on race to observe potential divergence in feature importance rankings across subgroups. A significant result for this hypothesis test indicates that the observed correlation between the rankings is unlikely to have arisen by chance. A high positive correlation suggests that the relative ordering of features is largely preserved. In contrast, failure to reject the null indicates that the rankings of feature importance differ. Importantly, significance does not necessarily imply a perfect match of rankings, nor does it measure the magnitude of changes in individual feature positions. Therefore, the test should be interpreted as evidence of whether the rankings are generally consistent or disrupted. Spearman's $\rho$ value is also reported to assess the practical extent of ranking changes.

\begin{table}[b!]
\centering
\caption{Predictive performance values of four models across the UTI, COMPAS, and AFib test datasets. The cell colors indicate changes in metrics following bias mitigation: green signifies an improvement, red indicates a decline, and white represents no change. $n$ is the number of samples in the test dataset, and $d$ is the number of features in the dataset. The F1 score is the harmonic mean of precision and recall. AUROC is the area under the receiver operating characteristic curve.
}
\begin{tabular}{llcccccc}
    \toprule
       & & \multicolumn{2}{c}{\makecell{UTI\\(n=387, d=7)}}  & \multicolumn{2}{c}{\makecell{COMPAS\\(n=2,037, d=9)}}  & \multicolumn{2}{c}{\makecell{AFib\\(n=15,959, d=210)}}\\
     \cmidrule(lr){3-8} 
     &  & Baseline & Mitigated & Baseline & Mitigated & Baseline & Mitigated \\
    \toprule
\multirow[t]{5}{*}{Logistic Regression} 
& Accuracy & 0.73 & \cellcolor{red!25}0.64  & 0.75 & \cellcolor{red!25}0.74 & 0.68 & 0.68\\
& F1 score & 0.56 & \cellcolor{red!25}0.51  & 0.70 & 0.70 & 0.69 & 0.69\\
& Precision & 0.66 & \cellcolor{red!25}0.50  & 0.74 & \cellcolor{red!25}0.73 & 0.67 & 0.67\\
& Recall & 0.49 & \cellcolor{green!25}0.52  & 0.66 & \cellcolor{green!25}0.67 & 0.70 & \cellcolor{green!25}0.71\\
& AUROC & 0.78 & \cellcolor{red!25}0.74  & 0.82 & \cellcolor{red!25}0.81 & 0.74 & \cellcolor{red!25}0.73\\
\midrule
\multirow[t]{5}{*}{Random Forest} 
& Accuracy & 0.71 & \cellcolor{red!25}0.65  & 0.74 & \cellcolor{red!25}0.72 & 0.99 & 0.99\\
& F1 score & 0.62 & \cellcolor{red!25}0.54  & 0.71 & \cellcolor{red!25}0.69 & 0.99 & 0.99\\
& Precision & 0.58 & \cellcolor{red!25}0.51  & 0.73 & \cellcolor{red!25}0.69 & 1.00 & 1.00\\
& Recall & 0.68 & \cellcolor{red!25}0.57  & 0.69 & 0.69 & 0.98 & 0.98\\
& AUROC & 0.78 & \cellcolor{red!25}0.76  & 0.82 & \cellcolor{red!25}0.81 & 0.99 & 0.99\\
\midrule
\multirow[t]{5}{*}{XGBoost} 
& Accuracy & 0.75 & \cellcolor{red!25}0.62  & 0.74 & \cellcolor{red!25}0.73 & 0.98 & 0.98\\
& F1 score & 0.57 & \cellcolor{red!25}0.42  & 0.70 & \cellcolor{red!25}0.69 & 0.98 & 0.98\\
& Precision & 0.74 & \cellcolor{red!25}0.47  & 0.75 & \cellcolor{red!25}0.71 & 0.99 & 0.99\\
& Recall & 0.46 & \cellcolor{red!25}0.38 & 0.64 & \cellcolor{green!25}0.67  & 0.97 & 0.97\\
& AUROC & 0.78 & \cellcolor{red!25}0.73  & 0.82 & \cellcolor{red!25}0.81 & 0.99 & 0.99\\
\midrule
\multirow[t]{5}{*}{Neural Network} 
& Accuracy & 0.75 & \cellcolor{red!25}0.64  & 0.74 & \cellcolor{red!25}0.65 & 0.99 & 0.99\\
& F1 score & 0.60 & \cellcolor{red!25}0.02  & 0.69 & \cellcolor{red!25}0.57 & 0.99 & 0.99\\
& Precision & 0.71 & \cellcolor{red!25}0.40  & 0.74 & \cellcolor{red!25}0.64 & 0.98 & \cellcolor{green!25}0.99\\
& Recall & 0.52 & \cellcolor{red!25}0.01  & 0.65 & \cellcolor{red!25}0.52 & 1.00 & 1.00\\
& AUROC & 0.77 & \cellcolor{red!25}0.65  & 0.82 & \cellcolor{red!25}0.79 & 1.00 & 1.00\\
    \bottomrule
\end{tabular}
\label{tab:metrics}
\end{table}

\section*{Results}

\textbf{Predictive Performance}. Table \ref{tab:metrics} displays the predictive performance metrics on test data for all models. All performance metrics, except the AUROC, were evaluated at a 50\% threshold. 

Across all three datasets, model performance unsurprisingly declined on average following mitigation. This pattern was most prevalent in the UTI dataset, where substantial decreases were observed across nearly all metrics and models. For example, in the neural network, UTI recall dropped sharply from 0.5290 to 0.0145, and the F1 score from 0.6083 to 0.0280, implying that only a network with poor predictive performance could satisfy the fairness constraint. Similar but less severe trends were observed in XGBoost and logistic regression models for UTI, with reductions in AUROC and precision. In contrast, models trained on the AFib dataset showed minimal performance decreases after mitigation. All metrics remained largely stable across all model types. Any differences noted were minimal, suggesting that fairness constraints could be incorporated in this context without strongly compromising predictive accuracy. For the COMPAS dataset, mitigation led to modest performance reductions. While AUROC and accuracy decreased slightly, particularly for the neural network and logistic regression models, F1 scores and recall remained relatively stable. Notably, random forest and XGBoost maintained most of their baseline performance post-mitigation.

Nearly all cases of performance improvement following mitigation manifested through increased recall. Since the fairness constraint is EOD, which enforces parity in both true positive and negative rates, this result suggests that the mitigation procedure primarily operated by increasing model sensitivity.

\begin{figure}[b!]
    \centering
    \includegraphics[width=\linewidth]{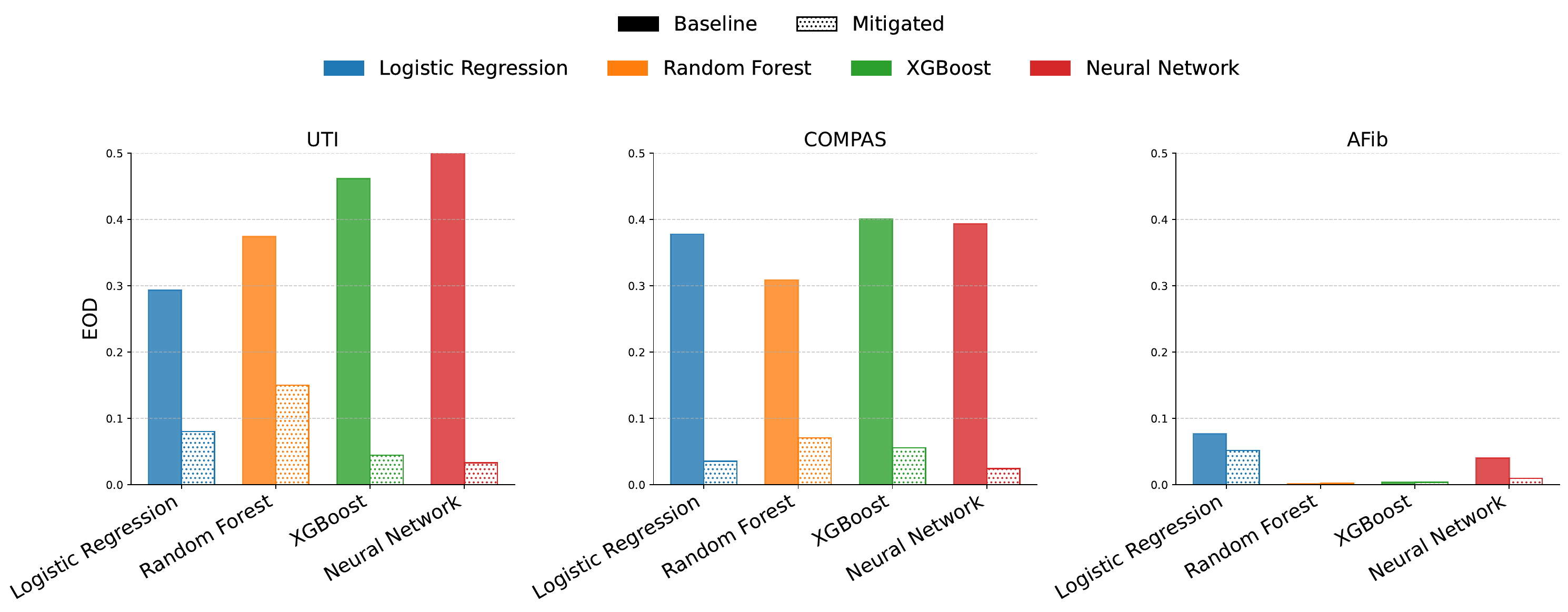}
    \caption{EOD values of four models evaluated on the UTI, COMPAS, and AFib test datasets before and after bias mitigation.}
    \label{fig:eod}
\end{figure}

\textbf{Fairness Performance}. To best interpret the changes in variable importance resulting from fairness mitigation, it is first necessary to assess the efficacy of the mitigation approach across each dataset and model combination. EOD for all models and datasets is presented in Figure \ref{fig:eod}. Lower EOD values are better.

For the UTI dataset, baseline EOD values start relatively high, ranging from 0.294 to 0.536, indicating that the initial models exhibited substantial group-level disparities. After fairness mitigation, the EOD values decrease sharply, ranging from 0.033 to 0.159. This considerable reduction demonstrates that the mitigation techniques were highly effective at improving fairness, especially for nonlinear models.

In contrast, the AFib dataset shows relatively low baseline EOD values, ranging from 0.002 to 0.077, suggesting that the data or model structure already supported relatively fair predictions before intervention. Nonetheless, mitigation further improves fairness, reducing EOD values to a range of 0.002 to 0.044. Notably, the tree-based methods achieve the lowest EOD values, indicating that these models may be better aligned with the structure of the AFib data. In contrast, the logistic regression and neural network models had higher levels of disparity both before and after mitigation.

For the COMPAS dataset, baseline EOD values are relatively high but consistent across models, ranging from 0.309 to 0.401. After mitigation, all models show a substantial and relatively uniform decrease in EOD, with post-mitigation values ranging from 0.0288 to 0.065. The consistent results in these findings align with prior literature highlighting the COMPAS dataset as a canonical example of embedded racial bias that can be mitigated.

\begin{figure}[p]
    \centering
     \hspace{-3cm} 
    \includegraphics[width=\textwidth]{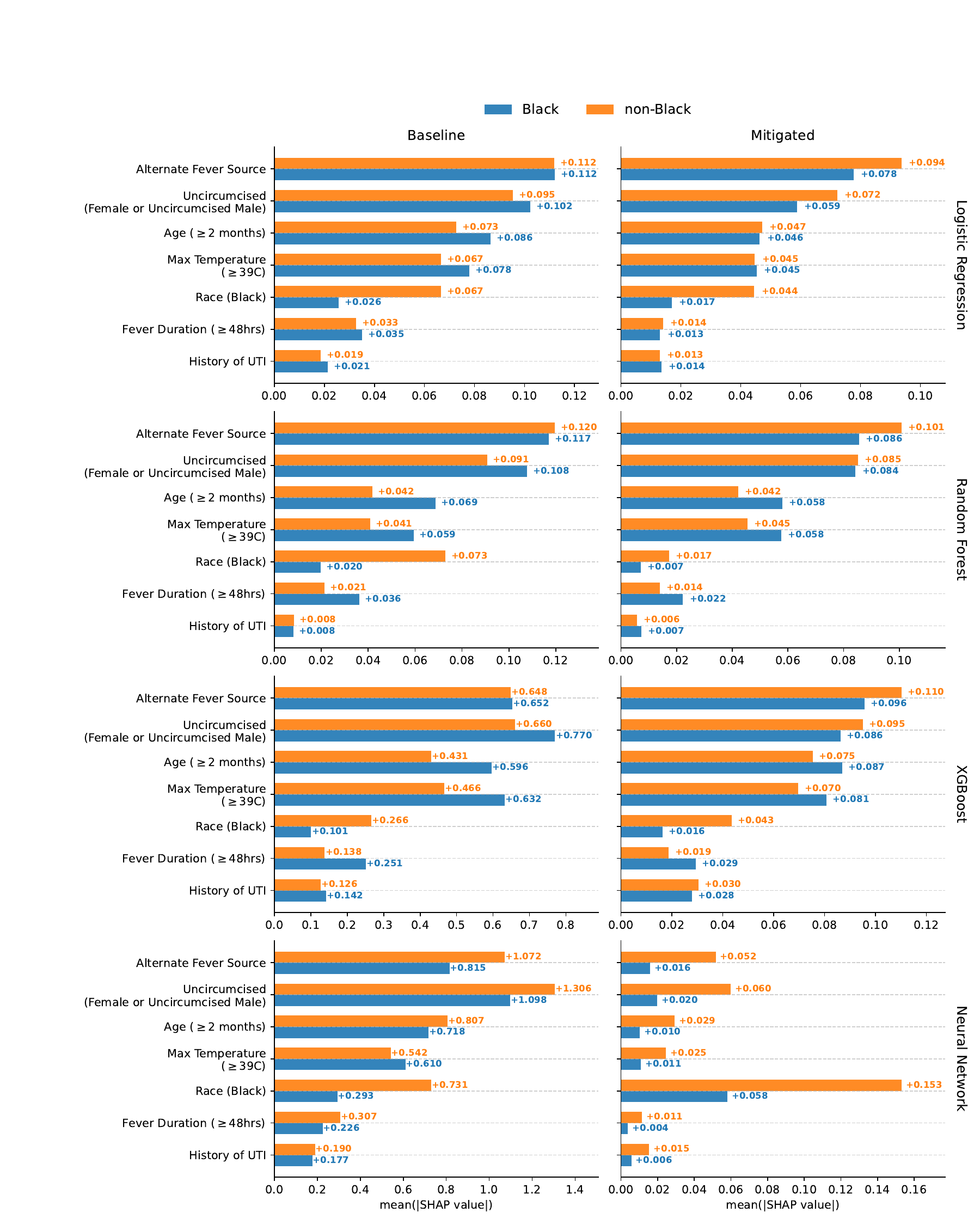}
    \caption{Feature importance of the seven features in the UTI dataset, as measured by global SHAP values, is shown for both baseline and bias mitigated models. Importance is computed separately for the two racial groups.}
    \label{fig:ranking}
\end{figure}

\textbf{Feature rankings.} An example of global explanations before and after mitigation can be seen in Figure \ref{fig:ranking}. This figure illustrates heterogeneity in the importance rankings across models and racial subgroups. For example, age is ranked third in importance overall in the baseline logistic regression across both racial groups, but fourth in the baseline XGBoost model across both groups. Table \ref{tab:spearman} presents the Spearman’s $\rho$ correlation values used to evaluate changes in SHAP-based feature rankings before and after bias mitigation across four models, three datasets, and two racial groups.

Complex models generally exhibit greater variability in feature importance rankings, whereas simpler models, such as logistic regression, show high stability. This variability is also not uniform across racial subgroups. For example, the random forest model for the UTI dataset showed stable feature rankings for Black patients, significantly different rankings for non-Black patients at 0.01, and overall differences at 0.001. Corresponding $\rho$ values indicate small changes for Black patients ($\rho = 0.96$) but relatively more changes for non-Black patients ($\rho = 0.86$). In the neural network model for COMPAS, rankings for Black patients differ only at the 0.001 significance level ($\rho = 0.80$). In contrast, rankings for non-Black patients differ at the 0.01 level ($\rho = 0.70$), reflecting greater instability in the non-Black group. Notably, models that exhibit failures to reject the null at the same significance level may still have variation in $\rho$ values. For example, the neural network for UTI shows a $\rho$ of 0.57 for Black patients, 0.75 for non-Black patients, and 0.61 overall, highlighting pronounced differences in feature importance stability across data samples. All tests for the AFib dataset yielded statistically significant results, even at the 0.001 level, indicating that rankings remained stable.

\vspace{5mm}
\begin{table}[ht]
\centering
\small
\caption{Spearman's $\rho$ correlation values used to assess changes in SHAP-based feature rankings before and after bias mitigation across four models, three datasets, and two racial groups. The symbols $\dag$ and $\ddag$ indicate that the feature rankings before and after bias mitigation differ significantly. Specifically, $\dag$ and $\ddag$ indicate a failure to reject the null hypothesis (that the feature rankings are uncorrelated) at the 0.01 and 0.001 significance levels, respectively.}
\begin{tabular}{llccc}
\toprule
 Model & Race & UTI & COMPAS & AFib\\
\midrule
   \multirow[t]{3}{*}{Logistic Regression}
      & Overall & 0.96 & 0.95  & 0.96\\
       & Black   & 0.89$^\ddag$ & 0.97  & 0.96\\
       & non-Black   & 0.96 & 0.90 & 0.95\\
   \multirow[t]{3}{*}{Random Forest}
      & Overall & 0.93$^\ddag$ & 0.85$^\ddag$  & 0.93\\
       & Black   & 0.96 & 0.72$^\dag$  & 0.94\\
       & non-Black   & 0.86$^\dag$ & 0.87$^\ddag$  & 0.84\\
   \multirow[t]{3}{*}{XGBoost}
      & Overall & 0.82$^\dag$ & 0.88$^\ddag$  & 0.95\\
       & Black   & 0.82$^\dag$ & 0.87$^\ddag$  & 0.95\\
       & non-Black   & 0.89$^\ddag$ & 0.80$^\ddag$  & 0.82\\
   \multirow[t]{3}{*}{Neural Network}
      & Overall & 0.61$^\dag$ & 0.67$^\dag$  & 0.91\\
       & Black   & 0.57$^\dag$ & 0.80$^\ddag$  & 0.92\\
       & non-Black   & 0.75$^\dag$ & 0.70$^\dag$  & 0.76\\
\bottomrule\\\\
\end{tabular}
\label{tab:spearman}
\end{table}

\section*{Discussion}
Across all three datasets, bias mitigation consistently reduces EOD values, confirming the generalizable effectiveness of exponentiated gradient reduction in improving fairness in model predictions. All models had some cost-to-performance trade-off to improve EOD, except for the XGBoost and Neural Network models trained on AFib, which did not exhibit a performance cost from mitigation. Among all performance metrics, recall improved most frequently, suggesting that EOD reduction was primarily achieved by inflating probabilities. The magnitude of this performance trade-off seems to be correlated with the dataset's sample size. UTI models saw the most significant performance drops, while the AFib models experienced minimal change.

The degree of EOD improvement varied across datasets and model types, underscoring that fairness mitigation is not uniformly effective across contexts. The interaction between model complexity, dataset characteristics, and inherent bias structure appears to influence how well fairness can be achieved. For example, tree-based models trained on AFib data demonstrated unbiased learning without mitigation and had a better EOD than the other mitigated models. This suggests that, in some cases, selecting a better-suited model architecture may be more optimal than relying on mitigating a current model. Moreover, the AFib models' lower baseline EOD could explain why no additional mitigation was necessary, whereas more disparate datasets, such as UTI, exhibit higher bias that requires greater intervention.

This contrast in fairness optimizations across architectures and datasets leads to divergent interpretations of the data, even when the predictive performance is similar. XGBoost exhibited comparable predictive performance but showed considerably different EODs for COMPAS before and after mitigation. Differences in feature attributions were observed, as shown in Table \ref{tab:spearman}. Substantial variation in $\rho$ values across race, as seen in the Neural Network for UTI, implies that the magnitude of change in feature importance rankings can differ across subgroups.

These results highlight a complex trade-off between performance, fairness, and explainability in ML. Applying bias mitigation can substantially alter feature importance rankings, especially in complex, nonlinear models. This raises important questions about the validity of explanations under such interventions. Achieving an optimal balance among these three axes is non-trivial and likely depends on factors such as data quantity, model architecture, and the specific application domain. It is important to note that this analysis is limited to three datasets and a single fairness mitigation technique. Future work should examine additional factors, including sensitivity to training and testing dataset sizes, alternative explainability methods (e.g., partial dependence, individual conditional expectation), and more fairness metrics. Overall, these findings underscore the need for continued investigation into the interactions between fairness and explainability to guide responsible deployment of ML in high-stakes domains.

\section*{Acknowledgements}
Research reported in this publication was supported by the National Institutes of Health under award number T15 LM007059 from the National Library of Medicine. It was also supported by a School of Computing and Information Predoctoral Fellowship to JWA.

\bibliographystyle{vancouver}
\bibliography{amia}  

\end{document}